\documentclass[sigconf,nonacm]{acmart}
\AtBeginDocument{%
  }

\acmISBN{978-1-4503-XXXX-X/2018/06}

\usepackage{subcaption}
\newcommand{\eg}{e.g.}
\begin{document}

\title{Towards Realistic 3D Emission Materials: Dataset, Baseline, and Evaluation for Emission Texture Generation}

\author{%
\textbf{Zhiyuan Zhang}$^{1,3}$
\textbf{Zijian Zhou}$^{2}$
\textbf{Linjun Li}$^{3}$
\textbf{Long Chen}$^{3}$
\textbf{Hao Tang}$^{1 \dagger}$
\textbf{Yichen Gong}$^{3 \ddagger}$\\[0.5em]
$^{1}$ Peking University \hspace{0.5em}
$^{2}$ King's College London \hspace{0.5em}
$^{3}$ XG Tech \hspace{0.5em}\\
$^{\dagger}$Corresponding Author \hspace{0.5em}
$^{\ddagger}$Project Lead\\
$^{\dagger}$bjdxtanghao@gmail.com \hspace{2em}
$^{\ddagger}$yichen.gong@xg.auto
}


\begin{abstract}
3D texture generation is receiving increasing attention, as it enables the creation of realistic and aesthetic texture materials for untextured 3D meshes. However, existing 3D texture generation methods are limited to producing only a few types of non-emissive PBR materials (\eg, albedo, metallic maps and roughness maps), making them difficult to replicate highly popular styles, such as cyberpunk, failing to achieve effects like realistic LED emissions. To address this limitation, we propose a novel task, emission texture generation, which enables the synthesized 3D objects to faithfully reproduce the emission materials from input reference images. Our key contributions include: first, We construct the Objaverse-Emission dataset, the first dataset that contains 40k 3D assets with high-quality emission materials. Second, we propose EmissionGen, a novel baseline for the emission texture generation task. Third, we define detailed evaluation metrics for the emission texture generation task. Our results demonstrate significant potential for future industrial applications. Dataset will be available at https://github.com/yx345kw/EmissionGen.
\end{abstract}


\begin{teaserfigure}
  \includegraphics[width=\textwidth]{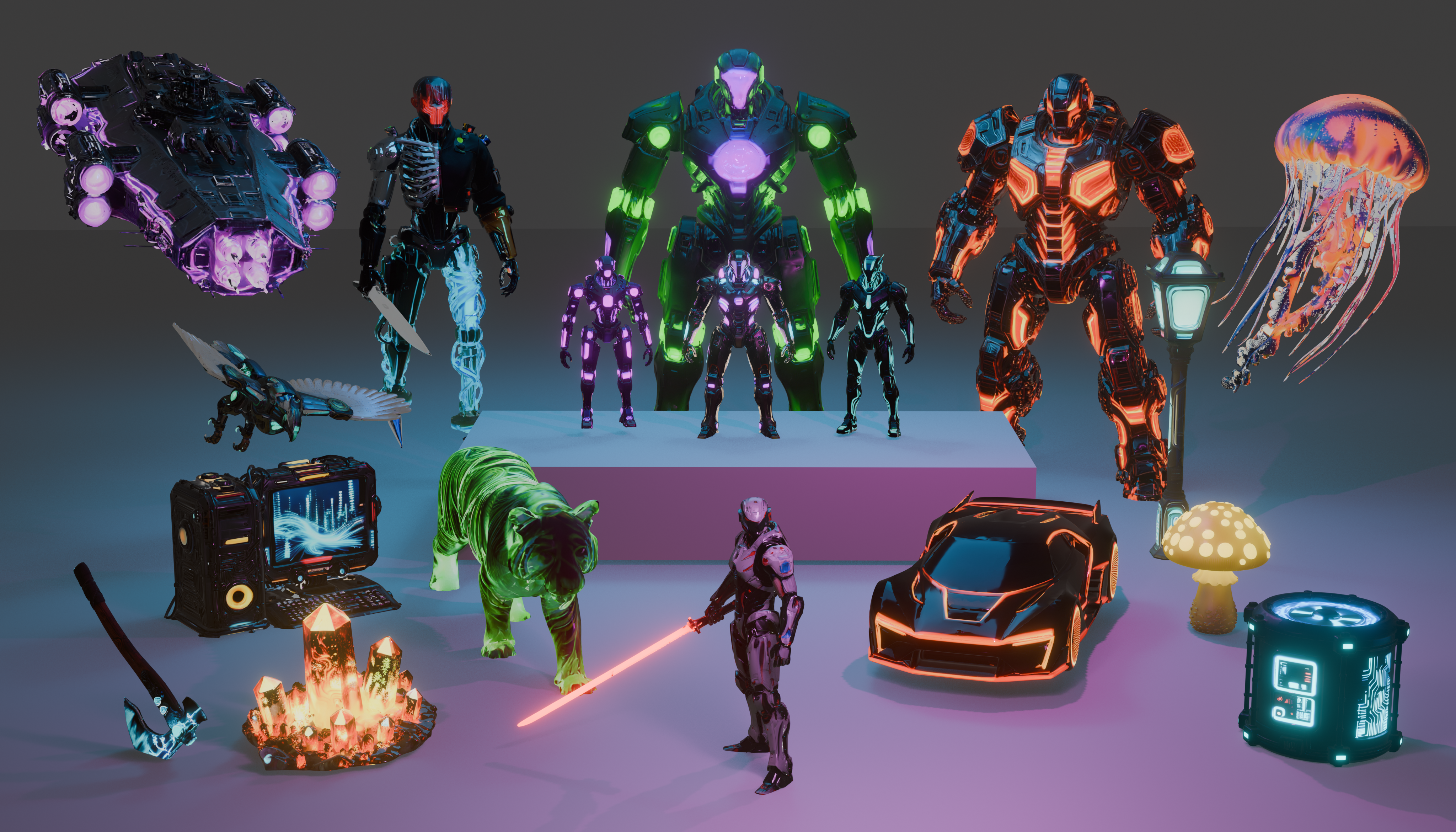}
  \caption{Gallery of our generated emission texture 3D assets.}
  \label{fig:teaser}
\end{teaserfigure}

\maketitle
\pagestyle{plain}

\section{Introduction}

Recent advances in 3D texture generation~\cite{hunyuan3d2025hunyuan3d,wang2025mage} have significantly expanded the ability to create realistic and visually appealing materials for colorless 3D geometries. Modern methods can now synthesize physically based rendering (PBR) materials, which disentangle intrinsic surface properties (e.g., albedo, normal, roughness) from external illumination effects. This decomposition enables accurate modeling of light–surface interactions, producing natural and physically consistent appearances that align with industrial rendering pipelines used in games, films, and digital twins.

Despite notable advances, current PBR texture generation models remain constrained in scope. They typically synthesize only a limited set of material maps, most commonly albedo, metallic, and roughness. As a result, these models cannot yet capture the full richness and stylistic diversity demanded by modern 3D content creation. For instance, cyberpunk-style assets with self-illuminating, LED-like glare effects, which are prevalent in digital art and game design, remain beyond the capabilities of existing systems. Although emission materials are widely used in physically based rendering and 3D content creation, existing PBR texture generation methods have paid little specific attention to them.

To bridge this gap, we introduce a novel task: emission texture generation. The goal is to synthesize textures that explicitly include an emission map, enabling 3D assets to exhibit self-emissive behaviors such as glowing lights, holographic panels, or neon signs. Incorporating emission textures extends the expressive power of generative 3D pipelines beyond passive reflection, supporting creative use cases in cinematic rendering, AR/VR asset design, and digital content creation.

A major obstacle to this task is the absence of high-quality, emission-aware datasets. As highlighted by StepFun3D~\cite{li2025step1x} and other prior works, the quality, diversity, and physical consistency of datasets are crucial for achieving robust texture generalization. To address this limitation, we construct Objaverse-Emission, the first large-scale dataset dedicated to emission texture generation. We first remove a large number of of low-quality and trivial small objects from Objaverse-1.0~\cite{deitke2023objaverse} and Objaverse-XL~\cite{deitke2023objaversexl}. Then we use Blender to obtain 3D assets that contain emission materials. We render the curated data using blender and then utilize luminous area statistics of each object to remove inappropriate data and get suitable emission strength label of each 3D asset. To ensure every 3D asset has only one object, we use VLM to judge all 3D assets. We end up with 40K 3D assets with physically consistent emission materials, exhibiting high visual diversity, a wide range of emission strengths, and rich semantic coverage.

Leveraging the Objaverse-Emission dataset, we train EmissionGen, a baseline model based upon state-of-the-art PBR texture generation architectures~\cite{hunyuan3d2025hunyuan3d,wang2025mage}. However, direct fine-tuning of existing models proves ineffective due to phenomena unique to emission materials. The first phenomenon we identify is \emph{glare disturbance}: light emitted from emissive regions illuminates surrounding non-emissive areas, causing the glare to be incorrectly baked into the albedo map. Accordingly, we introduce \emph{emission strength disentanglement training} that stabilizes the training and ensures a correct physical decomposition between emissive and non-emissive components. The second phenomenon is that emission maps are inherently sparse. Only emitting regions contain signal while the rest are black, making standard dense texture regression poorly suited for learning emission structure. To address this, we cast emission modeling as a binary segmentation problem and add an auxiliary \emph{emission-channel segmentation loss} with differentiable sigmoid-thresholded masks decoded from predicted and target clean latents.

To evaluate the newly defined emission texture generation problem, we also establish  a comprehensive benchmark, including metrics that capture both physical fidelity, perceptual realism and content diversity of emissive effects. Extensive experiments are conducted to analyze multiple baseline models under these metrics, revealing their respective strengths, limitations, and opportunities for improvement.

Our main contributions are summarized as follows:
\begin{itemize}
    \item \textbf{Novel Task Definition.} We introduce \emph{emission texture generation}, expanding existing texture synthesis tasks to include emission material modeling for the first time.
    \item \textbf{Dataset Construction.} Our curated Objaverse-Emission, a large-scale dataset of 40K high-quality 3D assets with emission materials, established a solid foundation for future research on emission texture generation.
    \item \textbf{Training Method and Baseline Model.} We identify two key challenges in emission texture generation. The first challenge is \emph{the sparsity of emission maps} and we use emission-channel segmentation loss to deal with this problem. The second challenge is \emph{glare disturbance} and we propose the emission strength disentanglement training strategy to address the issue.
    \item \textbf{Evaluation Benchmark.} We design comprehensive metrics and a standardized evaluation protocol for emission texture generation, and benchmark multiple approaches to provide quantitative and qualitative comparisons.
\end{itemize}
\section{Related Work}

\noindent \textbf{Large-scale 3D Datasets.} Large-scale 3D datasets have a pivotal role in advancing both geometric reconstruction and material generation. Early benchmarks such as ShapeNet~\cite{chang2015shapenet} indexed over 3 million CAD models, with 220K of them categorized into 3,135 WordNet synsets, establishing a foundation for data-driven study around 3D models. The Objaverse ecosystem~\cite{deitke2023objaverse, deitke2023objaversexl, lin2025objaverse++} has since pushed this boundary dramatically: Objaverse-1.0 contains roughly 800K diverse assets, while Objaverse-XL scales to over 10M, making it one of the largest collections to date. However, while these datasets encompass a broad range of geometries and styles, the texture quality and material completeness vary substantially. Many assets include incomplete PBR channels or synthetic baked lighting, which can lead to artifacts when training generative models. Objaverse++~\cite{lin2025objaverse++} addresses this through curated quality annotations, yeilding 10K manually verified 3D objects with detailed metadata, including aesthetic quality scores, texture color categories, composition flags, and transparency attributes, yet emission materials remain largely underrepresented.
This limitation motivates the construction of our Objaverse-Emission dataset, which explicitly targets physically consistent emission materials.

\noindent \textbf{Multi-view Texture Generation.}
Multi-view texture generation aims to synthesize globally consistent and geometrically aligned textures for 3D surfaces. 
Early optimization-based methods, such as CLIPMesh~\cite{mohammad2022clip}, Text2Mesh~\cite{michel2022text2mesh}, leverage the representational power of pretrained image-text model CLIP~\cite{radford2021learning}. By differentiably rendering 3D objects into 2D views, these methods minimize the semantic discrepancy between the rendered images and the given text prompts to generate desired textures.
Text2Tex~\cite{chen2023text2tex} performs progressive, depth-conditioned inpainting across multiple views. The generated views are back-projected onto the 3D model to iteratively complete the texture map. A subsequent texture refinement stage is usually essential for achieving coherent results, as stretched and blurry artifacts frequently appear near stitching regions.
Recent works explicitly maintain texture coherence within the model architecture.
MVPaint~\cite{cheng2025mvpaint} introduces a multi-view generation framework that mitigates inconsistencies arising from independently conducted view generations.
TexGen~\cite{yu2024texgen} performs diffusion-based synthesis directly on the unfolded UV map, ensuring alignment across surface regions.
RomanTex~\cite{feng2025romantex} enforces consistency through cross-attention between multiple views and the reference image, through incorporating 3D-aware rotary positional embeddings to inject geometric positional information.

\noindent \textbf{PBR and Emission Texture Generation.} 
Physically based rendering (PBR) texture generation aims to recover or synthesize intrinsic surface properties that ensure natural and consistent appearance under diverse lighting conditions. The key PBR attributes include albedo (base color), which defines the intrinsic surface color; roughness, which characterizes micro-scale surface irregularities; and metallic, which specifies whether a surface exhibits metal-like reflectance or dielectric-like scattering. Together, these parameters enable accurate modeling of surface reflectance that remains physically consistent and lighting-independent.

Theoretically, all aforementioned 3D texture generation frameworks can be extended to PBR material prediction. Recent works have increasingly emphasized on PBR texture rather than plain RGB textures. MaterialMVP~\cite{he2025materialmvp} proposes a multi-view PBR diffusion framework that produces illumination-invariant albedo, roughness, and metallic maps, enforcing geometry–material alignment through consistency-regularized training. Hunyuan3D-2.1~\cite{hunyuan3d2025hunyuan3d}, MVPainter~\cite{shao2025mvpainter} and MAGE~\cite{wang2025mage} generate production-ready PBR materials from images, achieving high-quality albedo, roughness, and metallic reconstruction with strong geometric alignment.

\section{Objaverse-Emission Dataset}
\label{sec:dataset}

\subsection{Preliminaries: PBR and Emission Materials}

\noindent \textbf{PBR materials.} Physically Based Rendering (PBR) materials models the surface property and enables 3D appearance rendering in a physically grounded manner. Their behavior is governed by the Bidirectional Reflectance Distribution Function (BRDF)~\cite{burley2012physically}, which models how incoming light is reflected at a surface. A PBR material consists of several texture maps. The albedo (or base color) encodes the intrinsic color spectrum reflectance property, independent of illumination. The roughness value controls the micro-scale surface irregularity and determines whether reflections appear sharp or diffuse. The metallic property specifies whether the surface behaves like a conductor or a dielectric, influencing both its reflectance color and overall energy conservation. 

\noindent \textbf{Emission material.}
An emission material extends standard PBR materials with two components: an \textit{emission map}, which specifies the color of self-luminous regions, and an \textit{emission strength}, a global scalar that controls their brightness. Unlike albedo, which reflects incoming illumination, emission adds radiance directly into the scene. As a result, emitting regions illuminate nearby surfaces, producing a characteristic glare that depends on geometry, viewpoint, and environmental lighting. This glare presents a unique challenge: the light contributed by self-illumination can bleed onto surrounding areas and be incorrectly baked into the albedo. Properly accounting for this glare effect is therefore essential when constructing datasets and training emission texture generation models.

\subsection{Dataset Curation}

\begin{figure*}[t]
  \centering
  \includegraphics[width=\textwidth]{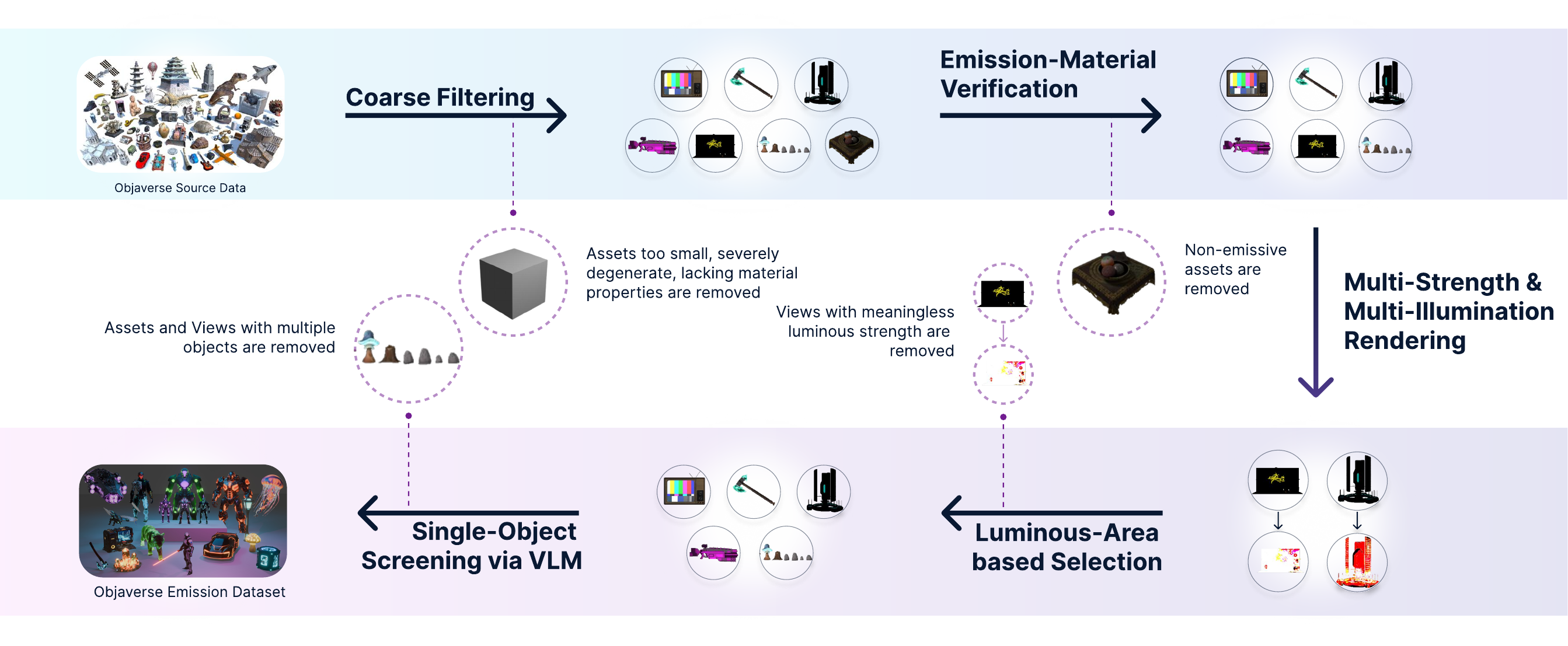}
  \caption{Overview of the Objaverse-Emission dataset construction pipeline. Starting from Objaverse~1.0 and Objaverse-XL, we perform (1) coarse asset filtering, (2) emission-material verification, (3) multi-strength and multi-illumination rendering, (4) luminous-area–based emission strength selection, and (5) VLM-based single-object screening. The resulting Objaverse-Emission dataset contains 40K high-quality 3D assets with reliable emission materials.}
  \label{fig:pipeline}
\end{figure*}

The Objaverse-Emission dataset is curated from Objaverse~1.0 and Objaverse-XL, combining high asset diversity with strict material-quality requirements. The pipeline consists of five stages, as shown in Figure \ref{fig:pipeline}.

\noindent\textbf{Step 1: Coarse Filtering.} Objaverse-XL contains many trivial or low-quality meshes, so we first remove assets that are too small, severely degenerate, or missing essential material information. For geometry coarse filtering, We measure the vertices number of each assets and discard the assets with low vertices number. Following \cite{shao2025mvpainter}, we compute the color entropy of texture maps of each object using Blender and use this value to filter assets with low-quality texture. In contrast, Objaverse~1.0 generally provides higher-quality geometry and materials, and we therefore retain all of its assets for subsequent processing.

\noindent\textbf{Step 2: Emission-Material Verification.} To verify the existence and validity of emission materials, we inspect each asset using Blender's Python API. We require that an asset have a valid emission map with a non-zero emission strength. To avoid trivial cases, we additionally ensure that the emission map is not identical to the base color map. To maintain compatibility with standard PBR texture generation frameworks and ensure that the filtered assets are of high quality, we exclude assets without metallic and roughness maps.

\noindent\textbf{Step 3: Emission \& Illumination-Diverse Rendering.} For each asset, emission strength interacts with geometry and lighting in complex ways: a high strength may remain visually plausible for some shapes but produce overwhelming glare for others. To capture this variability, we render every asset under a set of emission strength levels
$I_e = \{1.0, 1.25, 1.5, 1.75, 2.0, 2.5, 3.0\}$.

We further diversify the appearance by introducing randomized illumination conditions. Three types of illumination sources with randomized setups are applied to each asset:
(1) \textit{Point lights}, where three point sources are placed at random positions and their individual powers are randomized such that the total reaches approximately 180W;
(2) \textit{Area lights}, with randomly sampled radius, position, and power uniformly drawn from 10W to 500W; and
(3) \textit{Environment lights}, whose intensity is randomly chosen in the range $[0.5, 2.0]$.
These variations expose the asset to a broad spectrum of lighting behaviors, which is essential for understanding the interaction between emission and illumination in the environment.

For each lighting configuration, we capture ten canonical views: a top view, a bottom view, four orthogonal directions at zero elevation, and four diagonal directions at an elevation of $35.26^\circ$, corresponding to the cubical diagonal orientation. This multi-view, multi-emission, and multi-illumination rendering scheme comprehensively reflects the appearance variations induced by emission materials.

\noindent\textbf{Step 4: Luminous-Area–Based Selection.} Objects with excessively strong emission behave as interferences for stable training, as glare can dominate the rendered appearance. We therefore exclude both emission-dominated views and views in which emission is barely visible. Pixels with any emission channel exceeding 0.01 (normalized value range) are marked as the \textit{luminous area}. A luminous-area ratio below 0.01 indicates that the emission is nearly imperceptible, while a ratio above 0.8 suggests that the glare overwhelms the object and washes out its surface appearance. Emission strengths that produce views outside this admissible range are discarded. For each asset, the remaining strengths form an asset-specific valid set rather than a single label. This filtering ensures that only renderings with stable and visually meaningful emission behavior are retained.

\noindent\textbf{Step 5: Single-Object Screening via VLM.} Our generation model targets single-object scenarios, while many Objaverse assets contain multiple objects arranged within the same scene. To automatically filter out such cases, we employ the Vision–Language Model (VLM) Seed1.5-VL~\cite{guo2025seed1}. For each object, we feed its front, top, right, and left–rear–upper rendered views into the VLM, which then determines whether the preview contains a single object or multiple instances. We keep only single-object assets that possess valid emission materials.

\subsection{Diversity Analysis}

\begin{figure}[t]
    \centering
    \includegraphics[width=\linewidth]{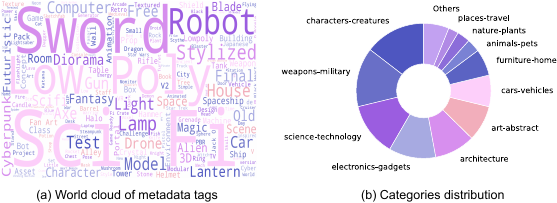}

  \caption{Dataset statistics of Objaverse-Emission: (a) word cloud of metadata tags; (b) categories distribution.}
  \label{fig:dataset_stats}
\end{figure}

We evaluate the diversity of Objaverse-Emission through multiple statistics, including metadata word clouds and category distributions. As illustrated in Figure~\ref{fig:dataset_stats}, the curated dataset spans a wide variety of shapes, materials, and emission behaviors, enabling robust learning for emission texture generation.

\begin{figure*}[t!]
    \centering
    \includegraphics[width=\textwidth]{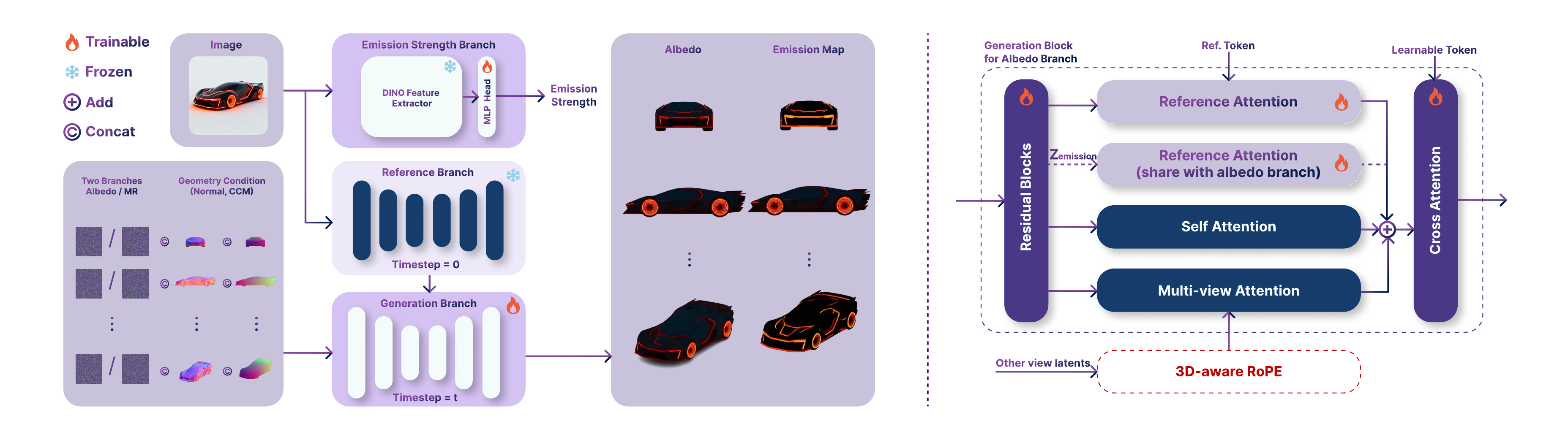}
    \caption{ Model architecture of EmissionGen. Following Hunyuan3D-2.1 Paint, we use a dual-branch architecture. The input to the reference branch and emission strength branch is a reference image. The input to the generation branch are concatenation of latent noise and multi-view normal maps and position maps of corresponding 3D mesh. The output of generation branch is multi-view albedo maps and emission maps. In each generation block, there are reference attention and multi-view attention block. For the alignment of albedo maps and emission maps, the reference attention of albedo branch is shared with emission branch.}
    \label{fig:model}
\end{figure*}

\section{EmissionGen}
\label{sec:emissiongen}

\subsection{Model Architecture}

The generation task is defined as estimating a global emission strength $I_e$ and synthesizing multi-view emission maps $v_i^{\text{emission}}$ together with their corresponding albedo maps $v_i^{\text{albedo}}$, given a reference image $\mathcal{I}_{\text{ref}}$ and an untextured mesh $\mathcal{M}$. Following recent multi-view generation paradigms, the model first produces a set of 2D projected maps from multiple viewpoints and then projects them back onto the mesh to cover the full 3D surface. Formally, the generation process $\mathcal{G}$ is defined as
\begin{equation}
\mathcal{G}(\mathcal{I}_{\text{ref}}, \mathcal{M}) \rightarrow \{I_e, v_i^{\text{emission}}, v_i^{\text{albedo}}\},
\label{eq:texture-gen}
\end{equation}
where $i$ indexes the set of rendered viewpoints.

\noindent\textbf{Model overview.}
Our method builds upon Hunyuan3D-2.1, which adopts a dual-branch architecture as shown in Figure~\ref{fig:model}. The \emph{reference branch} extracts feature representations from the reference image $\mathcal{I}_{\text{ref}} \in \mathbb{R}^{H \times W \times 3}$, while the \emph{generation branch} synthesizes multi-view emission and albedo maps. For each view $i$, its normal map $v_i^{\text{norm}} \in \mathbb{R}^{H \times W \times 3}$ and position map $v_i^{\text{pos}} \in \mathbb{R}^{H \times W \times 3}$ are encoded into latent space. The generation is performed in the latent space using a diffusion model:
\begin{equation}
  \{v^{\text{albedo}}_i, v^{\text{emission}}_i\} = EmissionGen(z^{\text{ref}}, z^{\text{norm}}_i, z^{\text{pos}}_i)
  \label{eq:emission-gen}
\end{equation}
where $z^{\text{ref}}$, $z^{\text{norm}}_i$, and $z^{\text{pos}}_i$ are the latent embeddings of the reference image, normal map, and position map, respectively.

\noindent\textbf{VAE encoding and decoding.}
Following the the latent diffusion approach, a variational autoencoder (VAE) encoder maps the input maps into the latent space:
\begin{equation}
v_i^{\text{pos}} \rightarrow z_i^{\text{pos}}, \qquad
v_i^{\text{norm}} \rightarrow z_i^{\text{norm}},
\end{equation}
while the VAE decoder reconstructs pixel-space outputs:
\begin{equation}
z_i^{\text{emission}} \rightarrow v_i^{\text{emission}}, \qquad
z_i^{\text{albedo}} \rightarrow v_i^{\text{albedo}}.
\end{equation}
This latent formulation substantially improves generation efficiency and training stability.

\noindent\textbf{Multi-view consistency and reference conditioning.}
To ensure cross-view coherence, the cross-view attention, as show in Figure~\ref{fig:model}, allows the model to exchange geometric and appearance cues across viewpoints. Furthermore, the reference image condition is also injected into the generation branch through cross-attention, guiding the generation process.

\noindent\textbf{Texture back-projection and emission strength estimation.}
To obtain the final textured 3D model, the generated multi-view emission maps are progressively projected and fused into UV space. For each rendering, a corresponding global emission strength is also required. We predict this scalar value using a lightweight MLP head composed of three fully connected layers, applied to features extracted from the reference image by DINOv2~\cite{oquab2023dinov2}.

\subsection{Emission-Channel Segmentation Loss}

Emission maps are inherently sparse. They only carry color at the region where emission material exists, whereas all non-emitting regions are black. This sparsity fundamentally differs from other material maps, which have defined values across the entire surface.

To address the sparsity problem of emission maps, we reformulate ``emitting vs.\ non-emitting'' as a binary segmentation task and add an additional segmentation loss on the emission channel of the model outputs. Following ~\cite{zhang2024pixel}, we decode the latent representation back to pixel space to compute our segmentation loss.

Our diffusion model adopts the v-prediction parameterization, where the network predicts a velocity target \(v\) rather than the noise \(\epsilon\). Let \(z_t\) denote the noisy latent at timestep~\(t\), obtained via the forward diffusion process \(z_t = \sqrt{\bar{\alpha}_t}\,z_0 + \sqrt{1-\bar{\alpha}_t}\,\epsilon\), where \(\epsilon \sim \mathcal{N}(0,I)\). The network \(v_\theta(z_t,t)\) predicts the velocity \(v = \sqrt{\bar{\alpha}_t}\,\epsilon - \sqrt{1-\bar{\alpha}_t}\,z_0\), and the ground-truth target velocity is computed analogously. The coefficients \(\sqrt{\bar{\alpha}_t}\) and \(\sqrt{1-\bar{\alpha}_t}\) are derived from the noise schedule and control the signal and noise levels, respectively. We recover clean-latent estimates from the predicted and target velocities:
\begin{equation}
\hat{z}_0^{\text{pred}} = \sqrt{\bar{\alpha}_t}\,z_t - \sqrt{1-\bar{\alpha}_t}\,v_\theta(z_t,t),
\end{equation}
\begin{equation}
\hat{z}_0^{\text{target}} = \sqrt{\bar{\alpha}_t}\,z_t - \sqrt{1-\bar{\alpha}_t}\,v.
\end{equation}

Both estimates are decoded to pixel space by the VAE decoder~\(D(\cdot)\). Ideally, the segmentation mask would be obtained by a hard threshold \(\mathbf{1}_{\{D(\hat{z}_0)>\tau\}}\), where \(\tau\) is the binarization threshold. However, the hard indicator function is non-differentiable, which prevents gradient back-propagation. We therefore adopt the sigmoid function as a differentiable surrogate:
\begin{equation}
\tilde{m}(x) = \sigma\!\bigl(k\,(D(x)-\tau)\bigr),
\label{eq:soft-mask}
\end{equation}
where \(\sigma(\cdot)\) is the sigmoid function, \(\tau\) is the binarization threshold, and \(k\) is a steepness parameter that controls how closely the smooth approximation resembles the hard step function (larger \(k\) yields a sharper transition). As \(k\to\infty\), \(\tilde{m}\) converges to the hard threshold mask \(\mathbf{1}_{\{D(x)>\tau\}}\).

The segmentation loss is then defined as
\begin{equation}
\mathcal{L}_{\text{seg}} := \mathbb{E}_{z_0,\,t}\;\text{DiceLoss}\!\left(\tilde{m}(\hat{z}_0^{\text{pred}}),\;\tilde{m}(\hat{z}_0^{\text{target}})\right),
\end{equation}
where DiceLoss~\cite{sudre2017generalised} measures the overlap between the predicted and target soft masks.

We combine the segmentation loss with the original multi-view consistent PBR texture generation loss \(\mathcal{L}_{\text{mcp}}\) via a weighting hyper-parameter \(\lambda=0.1\):
\begin{equation}
  \mathcal{L}_{\text{total}} := \mathcal{L}_{\text{mcp}} + \lambda \cdot \mathcal{L}_{\text{seg}}.
  \label{eq:total-loss}
\end{equation}

\subsection{Emission Strength Disentanglement Training}
\label{ssec:esd}

\begin{figure}[!htbp]
\centering

\begin{subfigure}{0.32\linewidth}
    \centering
    \includegraphics[width=\linewidth]{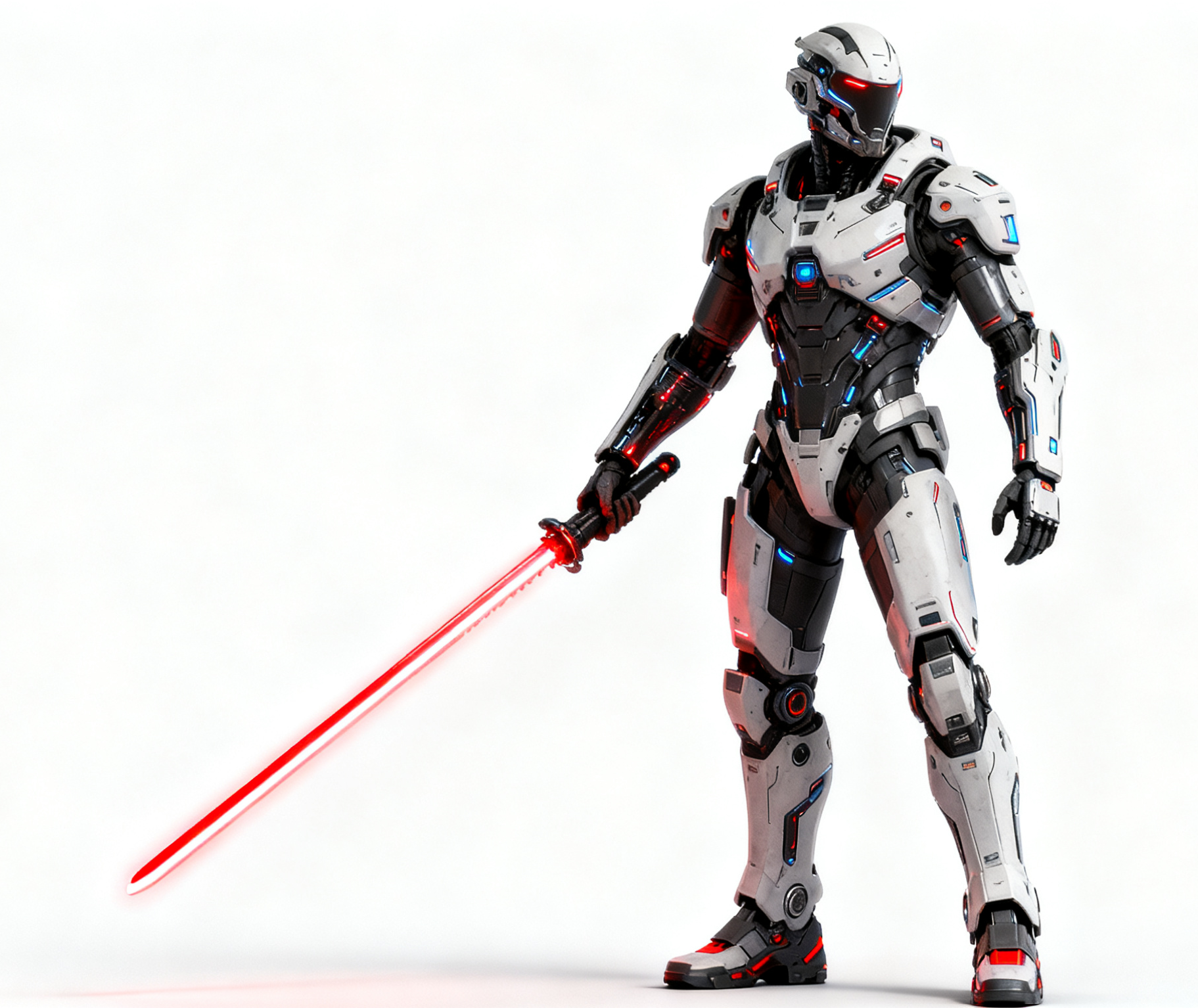}
  \end{subfigure}
  \hfill
  \begin{subfigure}{0.32\linewidth}
    \centering
    \includegraphics[width=\linewidth]{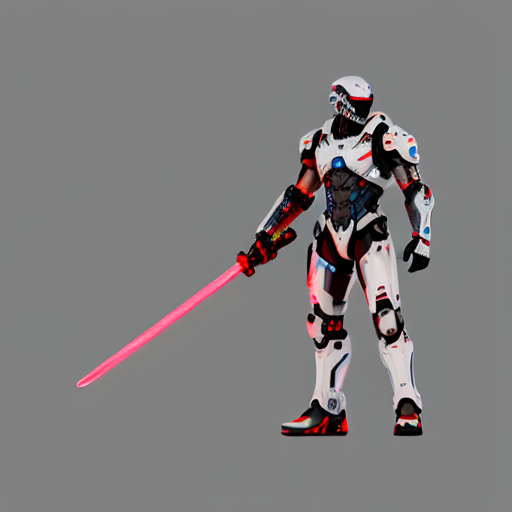}
  \end{subfigure}
  \hfill
  \begin{subfigure}{0.32\linewidth}
    \centering
    \includegraphics[width=\linewidth]{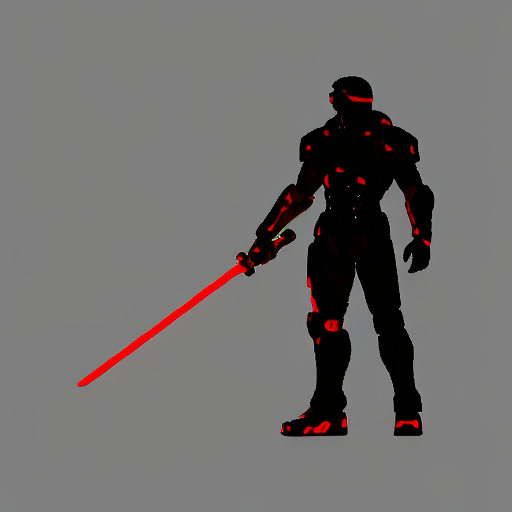}
  \end{subfigure}

\caption{Glare disturbance from emission materials. The left image shows the reference rendering, while the middle and right images depict the incorrectly generated albedo and emission maps. The strong glow emitted by the sword illuminates the adjacent lap region of the warrior, and this glare is mistakenly baked into both the albedo and emission maps during generation.}
\label{fig:emission_glow}
\end{figure}

Emission materials often illuminate nearby non-emissive regions, introducing glare disturbance, as illustrated in Figure~\ref{fig:emission_glow}. During texture generation, this effect leads to the unintended baking of glare into the albedo map, causing the emissive appearance to be incorrectly embedded in the object’s intrinsic color. Meanwhile, for emission material, when the emission strength is larger and cause more glare in the reference images, the intrinsic emission maps is unchanged. The model must identify the intrinsic property of emission maps and remain unaffected by the glare in the reference image.

\begin{figure}[!htbp]
    \centering
    \includegraphics[width=\linewidth]{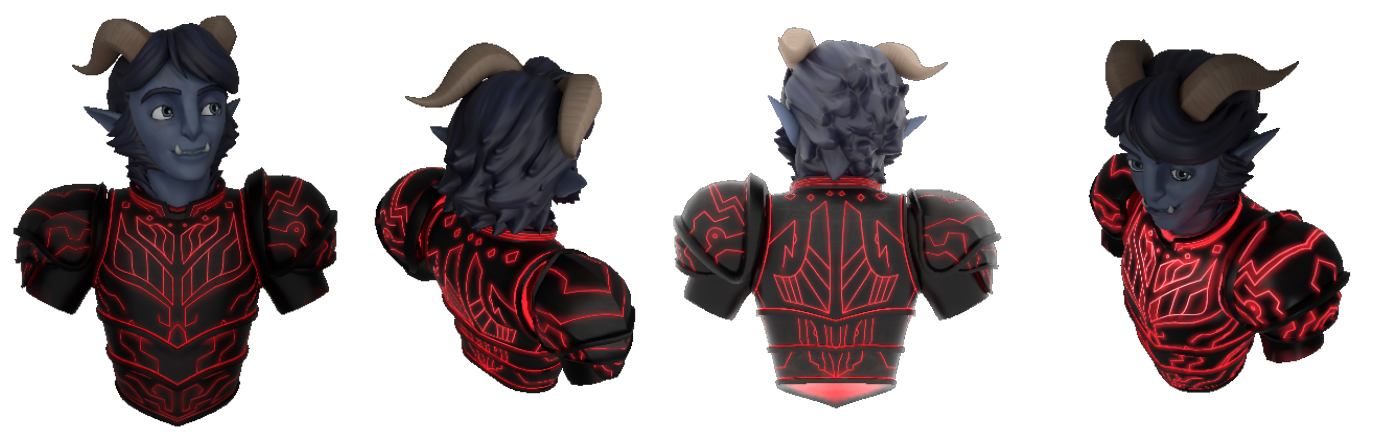}

  \caption{Multi-view renderings of a 3D asset under varying emission strengths and environmental illuminations. During emission-strength disentanglement training, reference images are randomly sampled across different viewpoints, lighting conditions, and emission intensities to encourage illumination-invariant material learning.}
  \label{fig:disentangle}
\end{figure}

To address this issue, we render each object under multiple controlled emission-strength settings. In the resulting training set, the reference images of a given object span a wide range of emission intensities and environmental lighting conditions, while the corresponding target albedo and emission maps remain fixed. Details of the lighting setup are provided in Section~\ref{sec:dataset}. During training, we randomly sample reference images with varying emission strengths and illumination conditions, as illustrated in Figure~\ref{fig:disentangle}. By exposing the model to both subtle and pronounced glare cases, it learns the interaction between emissive regions and their illumination effects on nearby surfaces. Consequently, the model becomes capable of correctly disentangling the intrinsic albedo and emission from the emission-induced glare.

\section{Experiments}
\begin{table*}[!htbp]
\centering
\caption{Quantitative evaluation on texture realism and diversity.}
\label{tab:texture_quant}
\captionsetup[subtable]{position=top,labelformat=empty}

\begin{subtable}[t]{0.48\linewidth}
\centering
\subcaption{Albedo Maps}
\label{tab:texture_quant_partA}
\resizebox{\linewidth}{!}{%
\begin{tabular}{lcccccc}
\toprule[1.5pt]
Method & PSNR${\uparrow}$ & SSIM${\uparrow}$ & LPIPS${\downarrow}$ & CLIP-FID${\downarrow}$ & FID${\downarrow}$ & CLIP-I${\uparrow}$ \\
\midrule
\textbf{EmissionGen} & \textbf{13.47} & \textbf{0.879} & \textbf{0.118} & \textbf{38.64} & \textbf{226.9} & \textbf{0.845} \\
MAGE-Emission & 11.46 & 0.823 & 0.134 & 80.25 & 280.7 & 0.821 \\
MVPainter-Emission & 11.03 & 0.836 & 0.127 & 84.35 & 282.6 & 0.833 \\
\bottomrule[1.5pt]
\end{tabular}%
}
\end{subtable}\hfill%
\begin{subtable}[t]{0.48\linewidth}
\centering
\subcaption{Emission Maps}
\label{tab:texture_quant_partB}
\resizebox{\linewidth}{!}{%
\begin{tabular}{lcccccc}
\toprule[1.5pt]
Method & PSNR${\uparrow}$ & SSIM${\uparrow}$ & LPIPS${\downarrow}$ & CLIP-FID${\downarrow}$ & FID${\downarrow}$ & CLIP-I${\uparrow}$ \\
\midrule
\textbf{EmissionGen} & \textbf{15.36} & \textbf{0.912} & \textbf{0.086} & \textbf{23.49} & \textbf{134.3} & \textbf{0.824} \\
MAGE-Emission & 13.06 & 0.871 & 0.095 & 43.29 & 264.3 & 0.786 \\
MVPainter-Emission & 13.27 & 0.878 & 0.091 & 44.59 & 265.8 & 0.793 \\
\bottomrule[1.5pt]
\end{tabular}%
}
\end{subtable}
\end{table*}
In this section, we first describe our baselines and evaluation metrics. We then present quantitative comparisons across different baselines, followed by qualitative results and ablation studies.

\subsection{Experiment setup}

We evaluate three 3D texture generation frameworks on our Objaverse-Emission dataset, while explicitly accounting for their different architectural assumptions. EmissionGen, introduced in Section~\ref{sec:emissiongen}, follows our target task definition and takes a reference image together with an untextured mesh as input. MVPainter~\cite{shao2025mvpainter} also satisfies this input setting, but adopts a two-stage pipeline composed of a novel-view synthesis stage and a PBR texture generation stage. We further evaluate MAGE~\cite{wang2025mage}, whose texture generator predicts PBR material maps for each synthesized view independently. However, unlike EmissionGen and MVPainter, the original first stage of MAGE is conditioned only on a reference image, rather than on both the reference image and the input mesh.

We initialize EmissionGen from the Hunyuan3D-2.1 Paint checkpoint~\cite{hunyuan3d2025hunyuan3d} and train it using the AdamW optimizer with a batch size of 8 and a learning rate of $5\times10^{-5}$. The model is trained for 50k steps on Objaverse-Emission. The input view number is 6 and the input resolution is 512$\times$512.

MVPainter~\cite{shao2025mvpainter} is a two-stage framework. Its first stage performs novel-view synthesis from the reference image together with mesh-derived multi-view normal maps and depth maps, and its second stage predicts multi-view PBR textures from the synthesized multi-view images. We fine-tune only the second-stage PBR texture generation model on Objaverse-Emission for 50k steps from the official checkpoint, obtaining MVPainter-Emission. Since its first stage is mesh-conditioned, MVPainter-Emission can be evaluated as a complete pipeline under the same input modality as EmissionGen.

MAGE~\cite{wang2025mage} is also a two-stage framework: the first stage synthesizes novel views from the reference image, and the second stage predicts albedo, normal, depth, metallic, and roughness maps for each synthesized view. However, the original first stage of MAGE does not take the input mesh as a condition, which is inconsistent with our task definition and yields synthesized views that are not well aligned with the target geometry to be textured. To make the comparison meaningful under our input setting, we therefore evaluate MAGE only as a texture-generation back-end: we replace its original first stage with the mesh-conditioned NVS stage of MVPainter, and feed the resulting geometry-aligned multi-view images into the second-stage texture generator of MAGE. Following the original paper~\cite{wang2025mage}, we fine-tune only the second-stage PBR generator, replacing the normal output channel with an emission channel. The model is initialized from the MAGE checkpoint and fine-tuned on Objaverse-Emission for 30k steps, yielding MAGE-Emission. We emphasize that this setting is not intended as a strictly full-parameter model comparison; instead, it assesses whether the MAGE texture generator remains competitive when all methods are supplied with mesh-compatible multi-view observations.

To train our emission strength estimation model, each reference image is paired with the exact emission strength used for its rendering in Objaverse-Emission; this scalar serves as the ground-truth label for the estimation head. The head is optimized with an MSE loss.

\subsection{Quantitative Comparison}

Quantitative evaluation of generated 3D emissive assets is ultimately carried out on rendered images. Commonly used fidelity and perceptual metrics are defined in 2D image space and therefore cannot be computed directly on a textured mesh representation. We therefore adopt a standardized render-then-evaluate protocol: for each method, we first obtain the final textured 3D asset, render it under matched camera viewpoints and illumination settings, and then compute image-space metrics on the resulting 2D views. This protocol is intended to assess the visual appearance of the generated 3D asset as perceived after rendering.

Under this protocol, we employ a comprehensive suite of metrics. For assessing the realism and fidelity of generated texture images, we report PSNR, SSIM~\cite{wang2004image}, LPIPS~\cite{zhang2018unreasonable}, and CLIP-I~\cite{radford2021learning}. For diversity assessment, we report FID~\cite{heusel2017gans} and CLIP-FID. For evaluating emission map accuracy, we use the Dice coefficient to measure the spatial accuracy of the predicted emission regions. Finally, to assess emission strength estimation, we report the RMSE between predicted emission strength and ground-truth emission strength associated with each sampled reference image.

We conduct quantitative evaluations using a 500-asset subset of Objaverse-Emission. EmissionGen performs the full texturing pipeline, generating albedo and emission maps, and ultimately a textured 3D asset directly from a reference image and an input mesh. MVPainter-Emission is also evaluated as a complete pipeline under the same input setting. For MAGE-Emission, we follow the protocol described above and combine the mesh-conditioned NVS stage of MVPainter with the fine-tuned second-stage texture generator of MAGE. Accordingly, its results should be interpreted as evaluating the MAGE texture back-end under a shared geometry-compatible front-end, rather than as a full-parameter model comparison against EmissionGen or MVPainter-Emission.

As shown in Table~\ref{tab:texture_quant} and Table~\ref{tab:emissive_region}, EmissionGen demonstrates clear advantages over MVPainter-Emission and MAGE-Emission. We attribute this improvement to the dual-branch reference image feature extraction design of EmissionGen and its reference-attention mechanism, which better suppress interference among different materials.

We conduct a user study on 15 people to compare EmissionGen, MVPainter-Emission, and MAGE-Emission in terms of quality and realism. EmissionGen achieved win rates of 55.5\% (albedo) and 53.1\% (emission). The results are consistent with the FID and CLIP-FID results, validating the reliability of our metrics.

The emission strength estimation achieves a RMSE of 0.4683, within a target range of 1.0 to 3.0. Since emission estimation prediction is a novel task without established baselines, we furthure conduct a user study on 15 people to validate its practical impact. We add noise with an RMSE of 0.4683 to the emission strength of test 3D assets and compare their quality against noise-free versions. 100\% participants found the difference imperceptible, demonstrating that this level of prediction error does not significantly affect visual quality.

\begin{figure*}[!htbp]
    \centering
    \includegraphics[width=\textwidth]{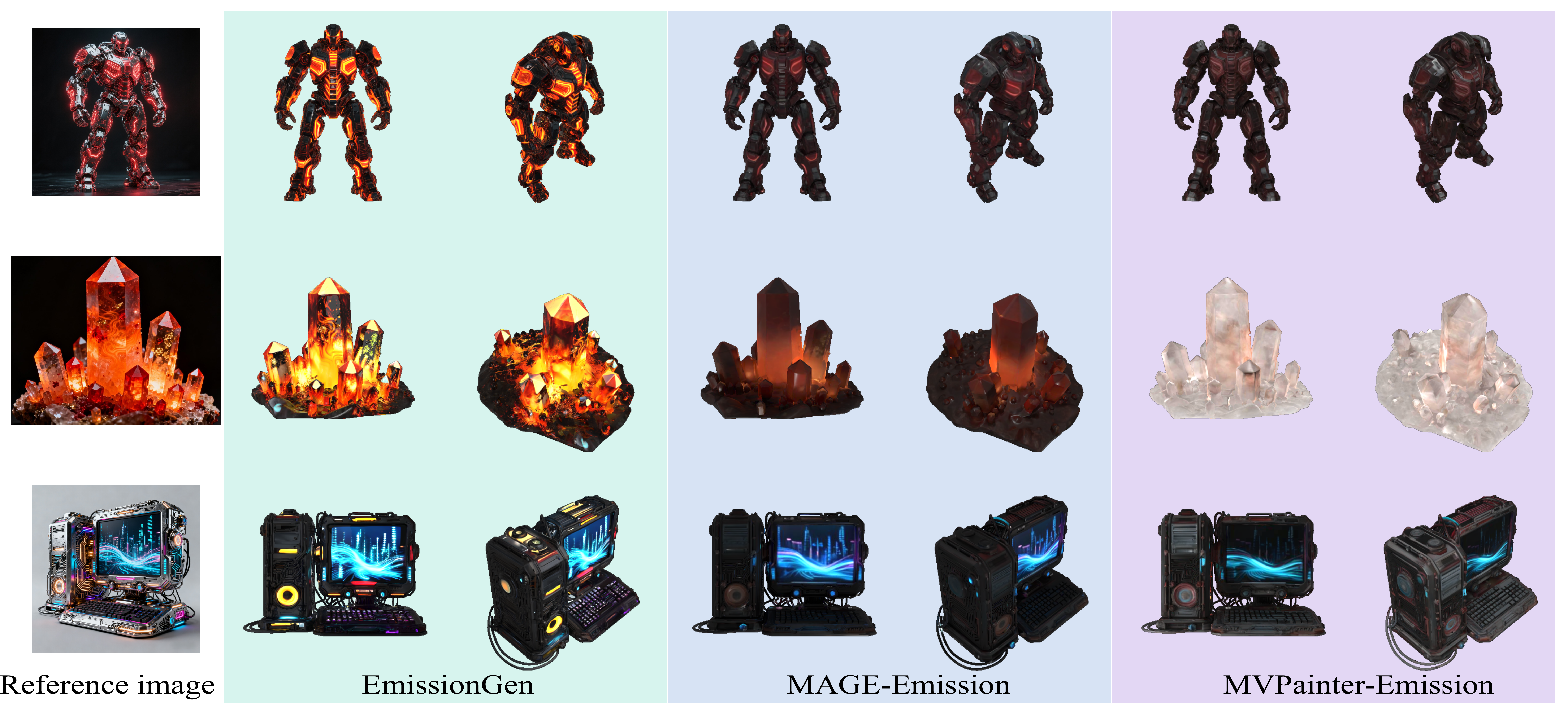}
    \caption{Qualitative comparison of different models. The left most column are reference images. On the right are multi-view images rendered from the complete synthesized textured 3D meshes generated by different models. The results show that EmissionGen achieves superior results compared to other methods. Please zoom in for better visualization of the details.}
    \label{fig:qualitative_results}
\end{figure*}

\begin{table}[h] 
\centering
\vspace{-8pt} 
\caption{Emission map accuracy.}
\label{tab:emissive_region}
\renewcommand{\arraystretch}{0.95} 
\begin{tabular}{lcc}
\toprule[1.2pt]
Method & Dice coefficient${\uparrow}$ \\
\midrule
\textbf{EmissionGen} & \textbf{0.98856} \\
MAGE-Emission        & 0.98721 \\
MVPainter-Emission   & 0.98749 \\
\bottomrule[1.2pt]
\end{tabular}
\vspace{-8pt} 
\end{table}

\subsection{Qualitative Results}

We also inspect qualitative results of EmissionGen, MVPainter-Emission, and MAGE-Emission. The input reference images are created by Doubao\cite{Volcengine2024Doubao},  HunyuanImage\cite{cao2025hunyuanimage} and Qwen-Image~\cite{wu2025qwen}. For the input untextured meshes, we use Hunyuan3D-3.0\cite{Hunyuan3D_website} Shape Generation to create them. EmissionGen takes a reference image and an untextured mesh as input and generates multi-view albedo maps and emission maps. The metallic and roughness maps are generated using the original Hunyuan3D-2.1 Paint, after which we apply the progressive projection algorithm following Hunyuan3D-2.1 to obtain the final textured mesh. MVPainter-Emission first synthesizes multi-view images from the reference image and the untextured mesh, and then generates multi-view PBR textures from the synthesized views, and finally using a similar projection algorithm to synthesis a complete textured mesh. For MAGE-Emission, consistent with our evaluation protocol, we feed the mesh-aligned views synthesized by the first stage of MVPainter into the second-stage PBR texture generator of MAGE to obtain the corresponding PBR textures, including the emission channel.

In Figure~\ref{fig:qualitative_results}, we showcase the textured 3D meshes produced by EmissionGen, MAGE-Emission and MVPainter-Emission. EmissionGen significantly outperforms than MAGE-Emission and MVPainter-Emission.

\subsection{Ablation Study}

\noindent \textbf{Ablation on Emission-channel Segmentatin Loss}. To validate the effectiveness of our emission-channel segmentation loss, we set segmentation loss weight to zero and train the model. As shown in Tables~\ref{tab:seg_loss_ablation}, the performance degrades when not using segmentation loss.

\begin{figure}[H]
  \centering
  \captionsetup[subfigure]{labelformat=empty}

  \begin{subfigure}{0.19\linewidth}
    \centering
    \includegraphics[width=\linewidth]{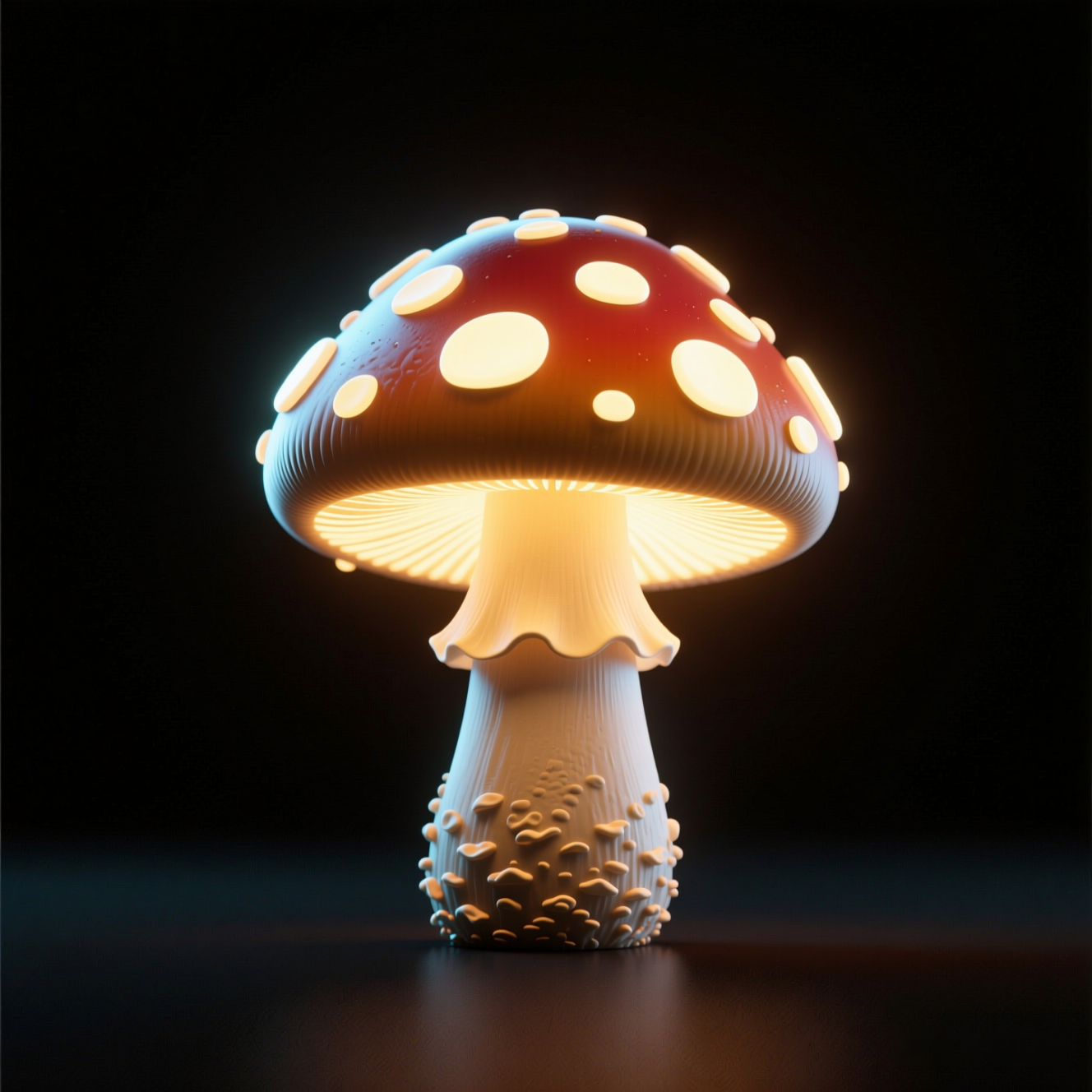}
    \caption{Reference image}
  \end{subfigure}
  \hfill
  \begin{subfigure}{0.19\linewidth}
    \centering
    \includegraphics[width=\linewidth]{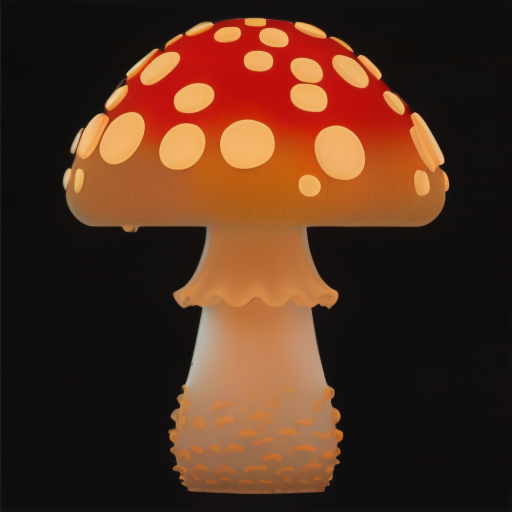}
    \caption{Albedo\\w/ ESDT}
  \end{subfigure}
  \hfill
  \begin{subfigure}{0.19\linewidth}
    \centering
    \includegraphics[width=\linewidth]{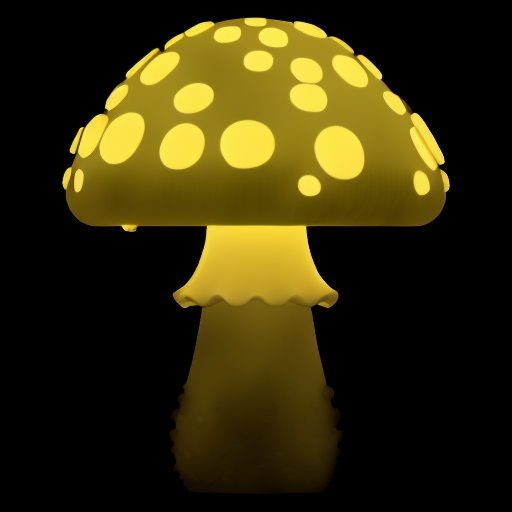}
    \caption{Emission\\w/ ESDT}
  \end{subfigure}
  \hfill
  \begin{subfigure}{0.19\linewidth}
    \centering
    \includegraphics[width=\linewidth]{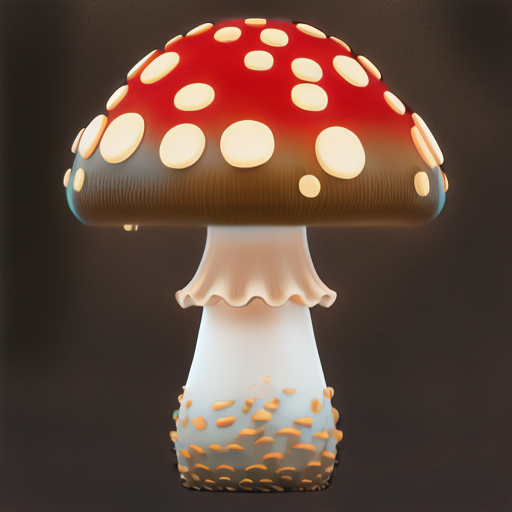}
    \caption{Albedo\\w/o ESDT}
  \end{subfigure}
  \hfill
  \begin{subfigure}{0.19\linewidth}
    \centering
    \includegraphics[width=\linewidth]{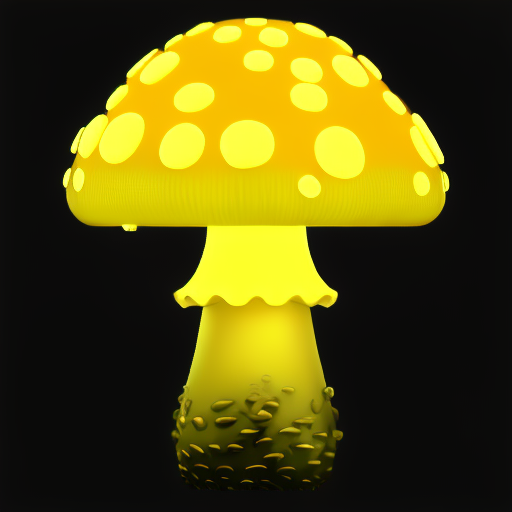}
    \caption{Emission\\w/o ESDT}
  \end{subfigure}

  \caption{With ESDT (Emission Strength Disentanglement Training), the glare caused by emission materials will not be baked into albedo maps, and the generated emission maps will be correct.}
  \label{fig:ablation_study}
\end{figure}

\noindent \textbf{Ablation on Emission-strength disentanglement strategy}. To validate the effectiveness of our emission-strength disentanglement strategy, we train a variant of the model using only images rendered with the default emission strength of 1.0. As shown in Figure~\ref{fig:ablation_study}, this model exhibits pronounced color leakage from emissive regions into non-emissive areas, affecting both the albedo and emission channels.
\begin{table}[t]
\centering
\caption{Emission-channel segmentation loss ablation study.}
\label{tab:seg_loss_ablation}
\renewcommand{\arraystretch}{0.9}
\resizebox{\columnwidth}{!}{
\begin{tabular}{lcccccc}
\toprule[1.2pt]
Method & PSNR${\uparrow}$ & SSIM${\uparrow}$ & LPIPS${\downarrow}$ & CLIP-FID${\downarrow}$ & FID${\downarrow}$ & CLIP-I${\uparrow}$ \\
\midrule
\textbf{w/ seg loss} & \textbf{15.36} & \textbf{0.912} & \textbf{0.086} & \textbf{23.491} & \textbf{134.3} & \textbf{0.824} \\
w/o seg loss & 14.43 & 0.891 & 0.088 & 24.245 & 139.1 & 0.810 \\
\bottomrule[1.2pt]
\end{tabular}
}
\vspace{-10pt} 
\end{table}

\section{Conclusion}

In this work, we introduced the new task of emission texture generation and curated Objaverse-Emission, a large-scale dataset of 40k 3D assets with high-quality emissive materials. Using this dataset, we developed EmissionGen as a strong baseline and established comprehensive evaluation metrics tailored for emission texture generation. To deal with the sparsity problem of emission maps, we use an emission-channel segmentation loss to enable the model to effectively accommodate to the unique structure patterns of emission maps. To address the challenge of glare interference, where emitted light contaminates surrounding regions, we proposed an emission-strength disentanglement training strategy that effectively separates intrinsic texture color from emissive intensity, preventing glare from being baked into the albedo and emission maps. Extensive experiments demonstrate that EmissionGen delivers high-quality, physically consistent emissive textures.

\bibliographystyle{ACM-Reference-Format}
\bibliography{sample-base}

\end{document}